\documentclass[10pt,conference,compsocconf]{IEEEtran}
\usepackage{times}
\usepackage{multirow}
\usepackage{array}
\newcolumntype{L}[1]{>{\raggedright\let\newline\\\arraybackslash\hspace{0pt}}m{#1}}
\newcolumntype{C}[1]{>{\centering\let\newline\\\arraybackslash\hspace{0pt}}m{#1}}
\newcolumntype{R}[1]{>{\raggedleft\let\newline\\\arraybackslash\hspace{0pt}}m{#1}}
\usepackage{graphicx}
\usepackage{caption}
\usepackage[table,xcdraw]{xcolor}
\usepackage{pgfplots}
\usepackage{tikz}
\usepackage{xcolor}
\usepackage[utf8]{inputenc}
\usepackage{kotex}

\ifCLASSINFOpdf
\else
\fi
\usepackage{amsmath}
\begin{document}
\pagenumbering{gobble}
%
\title{\textbf{\Large Squeezed Convolutional Variational AutoEncoder for \\[-1.5ex] Unsupervised Anomaly Detection in Edge Device Industrial Internet of Things}\\[0.2ex]}

\author{\IEEEauthorblockN{Dohyung Kim\IEEEauthorrefmark{1},
Hyochang Yang\IEEEauthorrefmark{1}\IEEEauthorrefmark{4},
Minki Chung\IEEEauthorrefmark{1}\IEEEauthorrefmark{4},
Sungzoon Cho\IEEEauthorrefmark{1}\IEEEauthorrefmark{6},
\\
Huijung Kim\IEEEauthorrefmark{3},
Minhee Kim\IEEEauthorrefmark{3},
Kyungwon Kim\IEEEauthorrefmark{3},
Eunseok Kim\IEEEauthorrefmark{3}
}
\IEEEauthorblockA{\IEEEauthorrefmark{1}Data Mining Center, Department of Industrial Engineering,
Seoul National University, Seoul, Republic of Korea}
\IEEEauthorblockA{\IEEEauthorrefmark{3}Data Analytics Lab, Software R\&D Center, Samsung Electronics Co. Ltd., Seoul, Republic of Korea}
\IEEEauthorblockA{Email: \{dohyung.kim, hyochang, minki.chung\}@dm.snu.ac.kr, zoon@snu.ac.kr, \\
\{friyahj.kim, mh0325.kim, kwkorea.kim, eunseok.kim\}@samsung.com}
\IEEEauthorblockA{\IEEEauthorrefmark{4} These authors contributed equally to the work, \IEEEauthorrefmark{6} Corresponding author}}


\maketitle

\begin{abstract}
In this paper, we propose Squeezed Convolutional Variational AutoEncoder (SCVAE) for anomaly detection in time series data for Edge Computing in Industrial Internet of Things (IIoT). The proposed model is applied to labeled time series data from UCI datasets for exact performance evaluation, and applied to real world data for indirect model performance comparison. In addition, by comparing the models before and after applying Fire Modules from SqueezeNet, we show that model size and inference times are reduced while similar levels of performance is maintained.

\end{abstract}



%
\IEEEpeerreviewmaketitle

\section{Introduction}
Manufacturing processes generate various sensor data simultaneously every time a product in produced. If the malfunction of a machine can be detected in advance, it is possible to perform predictive maintenance, which will reduce errors such as process malfunction and defective products. 

However, in manufacturing processes there are no labels for defects and malfunctions. Therefore, in such cases, a diagnosis of the behavior pattern of processes is necessary prior to predictive maintenance. In the absence of such labels, it is assumed that an abnormal behavior pattern is an anomaly.
 
Since sensor data are time series, it is important to find a model that reflects the characteristics of time series data. In order to do so, a Convolutional Neural Network (CNN) based Variational AutoEncoder (VAE) \cite{kingma2013auto} model for unsupervised anomaly detection was used.
 
Most systems in the manufacturing domain use a cloud-based approach with a central server to process data generated by local edge devices. However, lately edge-based approaches have been receiving more attention. Edge Computing in Industrial Internet of Things (IIoT) environments sends some of the computing load to edge devices for real time inference. In addition, data communication costs are very low compared to cloud-based methods, and reduces the burden on communication and computer infrastructure. However, neural networks that perform well in various way are compute and memory intensive, making it difficult to deploy them on edge devices with limited hardware resources. This problem was solved by using compressed neural networks.

In this paper, we propose a Squeezed Convolutional Variational AutoEncoder (SCVAE) model for unsupervised anomaly detection in time series sensor data for Edge Computing. First, the model performance was verified by applying SCVAE to labeled time series data. Next, for unlabeled real Computer Numerical Control (CNC) data, we propose the Match-General metric which was used to compare the performance of SCVAE with other models. Finally, model size and inference time, before and after compression, were compared to verify that SCVAE is a suitable model for use in edge devices.

\section{Related Work}

\subsection{Conventional unsupervised anomaly detection models}
Isolation Forest (IF) \cite{liu2008isolation} arbitrarily selects a feature and isolates the observation by randomly choosing the partition value between the maximum and minimum of the feature. The number of splits required to isolate each sample is the length from the root node to the leaf node and is used as a measure of anomaly. 
Local Outlier Factor (LOF) \cite{breunig2000lof} calculates the anomaly score by measuring the local density deviation for the neighborhood of a given sample. Samples whose local density deviations are much lower than their neighbors are classified as anomaly. 
One-Class SVM (OCSVM) \cite{scholkopf2001estimating} separates the data sample and origin from the kernel space and creates a complex boundary that describes normal data in the feature space. If a new sample is outside the decision boundary, it is classified as an anomaly. 
Elliptic Envelope (EE) \cite{rousseeuw1999fast} assumes that normal data is generated from a known distribution. It defines the shape of the distribution and defines points observed at the outline as anomalies. 

\subsection{Deep Learning based Unsupervised Anomaly detection models}
VAE is a generative model consisting of an Encoder and a Decoder. unsupervised anomaly detection using VAE was first proposed in \cite{an2015variational}. The probability that the actual data came from the output probability distribution of the encoder was defined as the reconstruction probability and was used to determine the anomaly in the data. Data samples with high reconstruction probability were considered normal, while data samples with low reconstruction probability were considered anomaly.

Generative Adversarial Networks (GAN) \cite{goodfellow2014generative} is a generative model consisting of a Discriminator and a Generator. Anomaly detection with Generative Adversarial Networks (AnoGAN) \cite{schlegl2017unsupervised} is an unsupervised anomaly detection model using GAN. After learning to generate a given data input, AnoGAN finds the noise z of the latent space that produces the most similar data to the given data and calculates the residual loss and the discrimination loss. The interpolated value of the two losses is used as the final loss, and this value is used as the anomaly score. However, it is not suitable for real time use because during inference it trains z to approximate the input X given by the Generator.

\subsection{Anomaly detection in time series data}
There is a study on using CNN to consider the relationship between different sensors and the time flow of time series manufacturing data of various sensors. In Fault Detection and Classification Convolutional Neural Networks (FDC-CNN) \cite{lee2017convolutional}, the CNN's receptive field was matched to the multivariate sensor signal, and the CNN filter was moved along the time axis to extract meaningful features from the sensor data.

\subsection{Edge Computing and model compression}
For Edge Computing, the size of the model should be small enough to be used in edge devices and should have a short inference time for real-time processing. In recent years, research has been conducted not only on the performance improvement of neural networks but also on reducing the size of the models while maintaining similar performance. Deep Compression \cite{han2015deep} reduced the size of the neural network by pruning and weight quantization, while SqueezeNet \cite{iandola2016squeezenet} and MobileNet \cite{howard2017mobilenets} used a modified CNN structure called Depthwise CNN and Fire Module (Squeeze and Expand layers), respectively to reduce the network size.

\section{Proposed Models}

\subsection{CNN-Variational Autoencoder (CNN-VAE)}
Since the input data is the time series of the sensors, it can be expressed in the form of two-dimensional image data. In this case, the time window size is set to 4, 8, and 16 according to the time set, and the dimension of the two-dimensional input data is time window (tw) $\times$ \# of features (\#f).

The input of the Encoder is the real data $x$ expressed as two-dimensional data and the output is the mean and variance parameters of the Gaussiam probability distribution of the latent variable $z$. By sampling from this probability distribution, the value of latent variable $z$ containing noise is obtained. The Encoder is modeled with a CNN structure to process the time series data. It consists of four convolutional layers and one fully connected layer.

The input of the Decoder is $z$ from the Encoder and the output is the mean and variance of the Gaussian probability distribution of the reconstructed $x$ from the latent variables. Like the Encoder, the Decoder was modeled using the CNN structure. It consists of five transpose convolutional layers and one fully connected layer. The detailed structure of the Encoder and Decoder is shown in Table \ref{CNC-VAE-archi}.

For all the convolutional layers and transpose convolutional layers in the Encoder and Decoder, the relu activation function \cite{nair2010rectified}, xavier initializer \cite{glorot2010understanding}, and batch normalization \cite{ioffe2015batch} with a decay of 0.9 were used. For the last fully connected layer, the activation function, weight initializer, and batch normalization were all excluded. Lastly, the size of the latent dimension ($\left|Z\right|$) was set to 100. 

\captionsetup{font={footnotesize,sc},justification=centering,labelsep=period}%
\begin{table}[htbp]

\centering
\renewcommand{\arraystretch}{1.3}
\begin{tabular}{C{1.5cm}C{1.0cm}C{0.8cm}C{0.8cm}C{1.0cm}}
\hline
\textit{Layer Type} & \textit{\# of outputs} & \textit{kernel size} & \textit{stride} & \textit{padding} \\
\hline
\multicolumn{5}{c}{\textit{Encoder}}\\[3pt]
\hline
conv & 16 & 3 & 1 & SAME \\ 
conv & 32 & 3 & 1 & SAME \\
conv & 64 & 3 & 1 & SAME \\
conv & 128 & 3 & 1 & VALID \\
fully connected & $\left|Z\right|$ $\times$ 2 & - & - & - \\
\hline
\multicolumn{5}{c}{\textit{Decoder}}\\[3pt]
\hline
trans conv & 128 & 3 & 1 & SAME \\
trans conv & 64 & 3 & 1 & SAME \\
trans conv & 32 & 3 & 1 & VALID \\
trans conv & 16 & 3 & 1 & VALID \\
trans conv & 1 & 3 & 1 & VALID \\
fully connected & tw $\times$ \#f $\times$ 2 & - & - & - \\
\hline
\end{tabular}
\caption{CNN-VAE architecture}\label{CNC-VAE-archi}
\end{table}
\captionsetup{font={footnotesize,rm},justification=centering,labelsep=period}%

In order to converge the parameters $\theta$ and $\phi$ of the Encoder and Decoder, the VAE trains in the direction of maximizing the Evidence Lower Bound (ELBO). ELBO includes the KL-divergence of the variational posterior and $z$ prior as well as the reconstruction log likelihood of $x$. The Adam optimizer \cite{kingma2014adam} was used for training. The formula for ELBO is shown in equation (\ref{eq:ELBO}).

\begin{equation}\label{eq:ELBO}
\resizebox{.9\hsize}{!}{$ELBO_i(\theta,\phi) = E_{q_{\theta}(z|x_i)}[\log p_\phi(x_i|z)] - KL(q_\theta(z|x_i)||p(z))$}
\end{equation}

The reconstruction probability of CNN-VAE indicates the probability of the data coming from the probability distribution of the given latent variable in the Encoder. The Anomaly score was set to $1-$ $recontruction$ $probability$. Data points with anomaly scores higher than threshold $\alpha$ are identified as anomalies and data points with anomaly scores lower than $\alpha$ are considered normal. The Anomaly score formula is as shown in equation (\ref{eq:as}).

\begin{equation}\label{eq:as}
Anomaly\:Score = 1-E_{q_{\theta}(z|x_i)}[p_\phi(x_i|z)]
\end{equation}

\captionsetup{font={footnotesize,sc},justification=centering,labelsep=period}%
\begin{figure}[htbp]
\centering%
 \includegraphics[width=9cm]{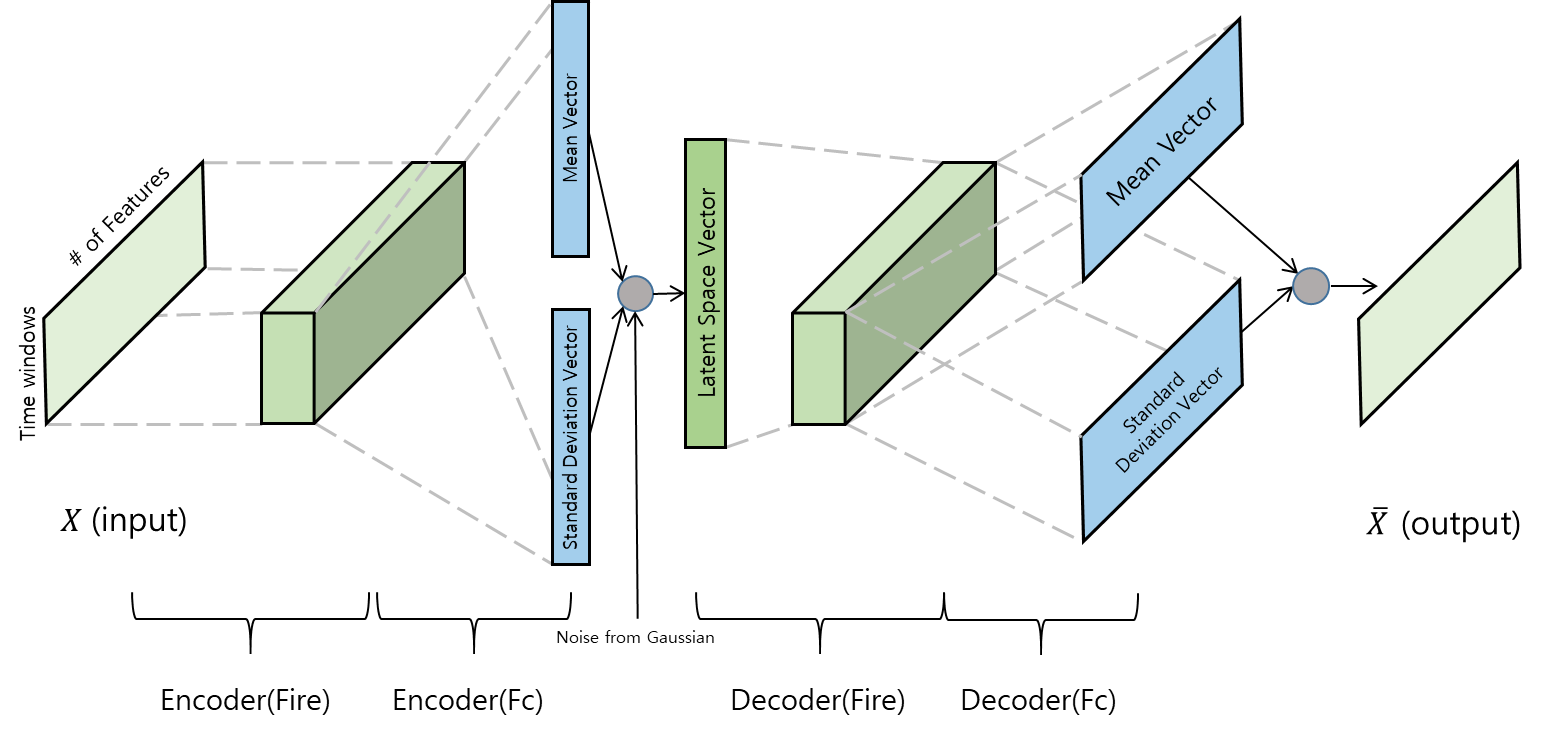}
\caption{SCVAE architecture}\label{SCVAE-figure}
\end{figure}
\captionsetup{font={footnotesize,rm},justification=centering,labelsep=period}%

\subsection{Squeezed Convolutional Variational Autoencoder (SCVAE)}
CNN-VAE shows good performance, but because of the large model size and slow inference it is inappropriate for use in edge devices where the computing resources are very limited. Therefore we propose the SCVAE model as shown in Figure \ref{SCVAE-figure}, a new model fit for use in Edge Computing. 
In SCVAE, the convolutional layers of CNN-VAE were converted into the form of Fire Modules used in SqueezeNet. A Fire Module consists of a squeeze layer and an extend layer. The squeeze layer consists of a 1 $\times$ 1 convolution filter, while the extend layer consists of one 1 $\times$ 1 convolution filter and one 3 $\times$ 3 convolution filter

The four convolutional layers used in the Encoder of CNN-VAE were replaced with one Fire Module, and the fire transpose convolutional layers used in the Decoder were replaced with one transpose Fire Module. The detailed structure of SCVAE is shown in Table \ref{SCVAE-archi}.

Relu activation, xavier initializer, and batch normalization with a decay of 0.9 were used in all of the convolutional and transpose convolutional layers included in the Fire Module and the transpose Fire Module. For the last fully connected layer, the activation function, weight initializer, and batch normalization were all excluded. The size of the latent dimension ($\left|Z\right|$) was set to 100. The loss function and Anomaly score are the same as for the CNN-VAE model. 

\captionsetup{font={footnotesize,sc},justification=centering,labelsep=period}%

\begin{table}[htbp]

\centering
\renewcommand{\arraystretch}{1.3}
\begin{tabular}{C{0.9cm}C{2.1cm}C{1.0cm}C{0.7cm}C{0.6cm}C{1.0cm}}
\hline
\multicolumn{2}{c}{\textit{Layer Type}} & \textit{\# of outputs} & \textit{kernel size} & \textit{stride} & \textit{padding} \\
\hline
\multicolumn{6}{c}{\textit{Encoder}}\\[3pt]
\hline
\multirow{3}{*}{fire} & squeeze / conv & 16 & 1 & 1 & SAME \\\
					  & extend1 / conv & 16 & 1 & 1 & SAME \\\
					  & extend2 / conv & 32 & 3 & 1 & SAME \\  
\multicolumn{2}{c}{fully connected} & $\left|Z\right|$ $\times$ 2 & - & - & - \\
\hline
\multicolumn{6}{c}{\textit{Decoder}}\\[3pt]
\hline
\multirow{3}{*}{trans fire} & squeeze / trans conv & 16 & 1 & 1 & SAME \\\
					  		& extend1 / trans conv & 16 & 1 & 1 & SAME \\\
					  		& extend2 / trans conv & 1 & 3 & 1 & SAME \\     
\multicolumn{2}{c}{fully connected} & tw $\times$ \#f $\times$ 2 & - & - & - \\
\hline
\end{tabular}
\caption{SCVAE architecture}\label{SCVAE-archi}
\end{table}
\captionsetup{font={footnotesize,rm},justification=centering,labelsep=period}%

\section{Experiment}
Experiments were conducted in three different directions. First, to verify the performance of SCVAE, comparison of SCVAE with baseline unsupervised anomaly detection algorithms, such as IF, LOF, OCSVM, and EE, was conducted with labeled time series data. Second, the evaluation metric of Match-General is presented to check the performance of SCVAE for unlabeled real world CNC data and indirectly compare it with other unsupervised anomaly detection algorithms. Third, the model size and inference time of SCVAE were compared with CNN-VAE to confirm that SCVAE is suitable for Edge Computing. Batch size of 64 and learning rate of 0.0002 were used in the experiment.

\subsection{SCVAE performance on labeled data}
The labeled time series data sets used are the one hour peak dataset in Ozone \cite{Kun2008ozone}, and Occupancy dataset \cite{Luis2016occupancy}, all from the UCI datasets. For comparison of SCVAE with IF, LOF, OCSVM, EE, the experiment was conducted by changing the time window from 4 to 8 to 16. In each dataset, there is an anomaly label for each row. If an anomaly exists in any row of the time window, the time window is regarded as anomaly. The descriptions for the above datasets are in Table \ref{timeseries-labeled-data}.

\captionsetup{font={footnotesize,sc},justification=centering,labelsep=period}%
\begin{table}[htbp]

\centering%
\renewcommand{\arraystretch}{1.3}
\begin{tabular}{C{1.5cm}C{1.5cm}C{1.5cm}C{1.5cm}}
\hline
\textit{Dataset} & \textit{\# of Samples} & \textit{\# of Features} & \textit{Anomaly Ratio($\%$)} \\
\hline
ozone & 2536 & 73 &  2.87 \\
occupancy & 8143 & 6 & 21.23 \\
\hline
\end{tabular}
\caption{Timeseries Labeled Dataset}\label{timeseries-labeled-data}
\end{table}
\captionsetup{font={footnotesize,rm},justification=centering,labelsep=period}

For the performance evaluation of the four existing algorithms and the proposed SCVAE, the Area Under the Precision-Recall Curve (PRAUC) was used. The results of this experiment are shown in Table \ref{timeseries-labeled-data-result}.

\captionsetup{font={footnotesize,sc},justification=centering,labelsep=period}%
\begin{table}[htbp]

\centering%
\renewcommand{\arraystretch}{1.3}
\begin{tabular}{C{1.6cm}C{0.8cm}C{0.8cm}C{1cm}C{0.8cm}C{0.8cm}}
\hline
\textit{Dataset\tiny{(Time Window)}} & \textit{IF} & \textit{LOF} & \textit{OCSVM}  & \textit{EE} & \textbf{\textit{SCVAE}} \\
\hline
ozone(4)             & 85.71             & 88.72                  & 88.18           & 86.21              & \textbf{95.35} \\
ozone(8)             & 83.02             & 83.27                  & 80.06           & 81.99              & \textbf{96.89} \\
ozone(16)            & 68.05             & 68.83                  & 64.70           & 69.26              & \textbf{81.78} \\
\hline
occupancy(4)   & 95.93             & 74.07                  & 80.99           & \textbf{99.17}     & 98.54          \\
occupancy(8)   & 96.81             & 74.89                  & 81.19           & 98.76              & \textbf{99.10} \\
occupancy(16)  & 96.62             & 75.57                  & 78.39           & 99.21              & \textbf{99.23} \\
\hline
\end{tabular}
\caption{PRAUC Comparison on Timeseries Labeled Dataset}\label{timeseries-labeled-data-result}
\end{table}
\captionsetup{font={footnotesize,rm},justification=centering,labelsep=period}%

For the ozone dataset, SCVAE overwhelms the performance of other non neural network algorithms in time windows 4, 8, and 16. For the occupancy dataset, the performance is better than the other non neural network algorithms for time window 8 and 16, but the performance is slightly lower than EE in time windows 4 . However, it can still be seen that the performance is significantly better than the other three non neural network algorithms. In conclusion, SCVAE is better than conventional anomaly detection algorithms in most cases. 

\subsection{SCVAE performance on CNC data using Match-General Metric} 
In the case of unlabeled CNC data, a metric called Match-General is proposed for performance evaluation. Performance comparison of SCVAE with the baseline unsupervised anomaly detection algorithms was done using this metric. In the Match-General metric, common labels are created and evaluated as actual labels. Anomaly in common labels are labeled as anomaly if the 3 or more of the 6 models identify the label as anomaly. The higher the value of Match-General, the better the performance. To evaluate for various types of CNC data, four types of CNC data were used. Each CNC data was arbitrarily designated as CNC A, B, C, D. The description of the CNC data is in Table \ref{cnc-data} and the results of this experiment are shown in Table \ref{CNC-data-result}. Considering CNC data characteristics, anomaly ratio was set to 5 percent.

\captionsetup{font={footnotesize,sc},justification=centering,labelsep=period}%
\begin{table}[htbp]

\centering%
\renewcommand{\arraystretch}{1.3}
\begin{tabular}{C{1.5cm}C{1.5cm}C{1.5cm}}
\hline
\textit{CNC} &\textit{\# of Samples} & \textit{\# of Features}  \\
\hline
A & 258697 & 31 \\
B & 310174 & 43 \\
C & 111770 & 43 \\
D & 602075 & 37 \\
\hline
\end{tabular}
\caption{CNC Dataset}\label{cnc-data}
\end{table}
\captionsetup{font={footnotesize,rm},justification=centering,labelsep=period}

\captionsetup{font={footnotesize,sc},justification=centering,labelsep=period}%
\begin{table}[htbp]

\centering%
\renewcommand{\arraystretch}{1.3}
\begin{tabular}{C{0.6cm}C{0.8cm}C{0.9cm}C{0.9cm}C{0.8cm}C{1cm}C{0.9cm}}
\hline
CNC   & IF              & LOF             & OCSVM  & EE              & CNNVAE          & SCVAE           \\
\hline
A(4)  & 0.9435          & \textbf{0.9565} & 0.4806 & 0.9532          & 0.9435          & 0.9371          \\
A(8)  & \textbf{0.9677} & 0.9323          & 0.6484 & 0.9645          & 0.9290          & 0.9355          \\
A(16) & 0.9516          & 0.9516          & 0.4532 & \textbf{0.9548} & 0.9355          & 0.9387          \\
\hline
B(4)  & 0.9413          & 0.9329          & 0.3208 & 0.9329          & \textbf{0.9665} & \textbf{0.9665} \\
B(8)  & 0.9266          & 0.9350          & 0.2180 & 0.9392          & \textbf{0.9686} & 0.9602          \\
B(16) & 0.9476          & 0.9476          & 0.3585 & 0.9434          & \textbf{0.9560} & \textbf{0.9560} \\
\hline
C(4)  & \textbf{0.9686} & 0.9507          & 0.5695 & 0.9327          & 0.9507          & 0.9327          \\
C(8)  & \textbf{0.9552} & 0.9462          & 0.8565 & 0.9462          & 0.9372          & \textbf{0.9552} \\
C(16) & 0.9507          & 0.9507          & 0.4215 & 0.9417          & 0.9507          & \textbf{0.9596} \\
\hline
D(4)  & 0.9444          & 0.9493          & 0.6221 & 0.9493          & \textbf{0.9510} & \textbf{0.9510} \\
D(8)  & 0.9535          & 0.9419          & 0.7467 & 0.9452          & 0.9485          & \textbf{0.9568} \\
D(16) & 0.9394          & 0.9277          & 0.1312 & 0.9377          & 0.9560          & \textbf{0.9626} \\              
\hline
\end{tabular}
\caption{Performance(MG) Comparison on CNC Dataset}\label{CNC-data-result}
\end{table}
\captionsetup{font={footnotesize,rm},justification=centering,labelsep=period}%

Table \ref{CNC-data-result} shows that the SCVAE has a high Match-General value for most CNC equipment. SCVAE showed best performance in CNC C and CNC D, and maintained high performance in CNC A and CNC B as well. 

\subsection{Comparison between SCVAE and CNN-VAE} 
Experiments were conducted to compare the learning time, inference time, and model size of squeezed model, SCVAE, with the full model, CNN-VAE. The PRAUC of both models were also compared to see whether the SCAVE model maintained performance levels. Figure \ref{fig:graph} and Table \ref{model-size-inference-time} show the experimental results for occupancy data with a time window of 16. Refer to the appendix for the results of the other datasets. 

\definecolor{bblue}{HTML}{4F81BD}
\definecolor{rred}{HTML}{C0504D}
\definecolor{ggreen}{HTML}{9BBB59}
\definecolor{ppurple}{HTML}{9F4C7C}
\captionsetup{font={footnotesize,sc},justification=centering,labelsep=period}%
\begin{figure}[htbp]
\centering%
\begin{tikzpicture}[trim left=-0.65cm]
    \begin{axis}[
        width  = 0.49*\textwidth,
        height = 4.5cm,
        major x tick style = transparent,
        ybar=2*\pgflinewidth,
        bar width=8pt,
        ymajorgrids = true,
        ylabel = {Ratio},
		ylabel near ticks,
        symbolic x coords={Learning, Inference, Size(Mb), PRAUC},
        xtick = data,
        scaled y ticks = false,
        enlarge x limits=0.28, 
        ymin=0,
        legend cell align=left,
        legend style={
                at={(0.5, 1.20)},
                anchor=north,
                legend columns=-1,
                column sep=1ex,
                draw=none
        },
        nodes near coords,
        nodes near coords align={vertical},
        axis y line*=none,
        axis x line*=none
    ]
        \addplot[style={bblue,fill=bblue,mark=none}]
            coordinates {(Learning, 0.67) (Inference, 0.68) (Size(Mb), 0.75) (PRAUC, 1)};

        \addplot[style={rred,fill=rred,mark=none}]
             coordinates {(Learning, 1) (Inference, 1) (Size(Mb), 1) (PRAUC, 1)};

        \legend{SCVAE, CNN-VAE}
    \end{axis}
\end{tikzpicture}
\caption{Comparison between SCVAE and CNN-VAE on occupancy(16) dataset}\label{fig:graph}
\end{figure}
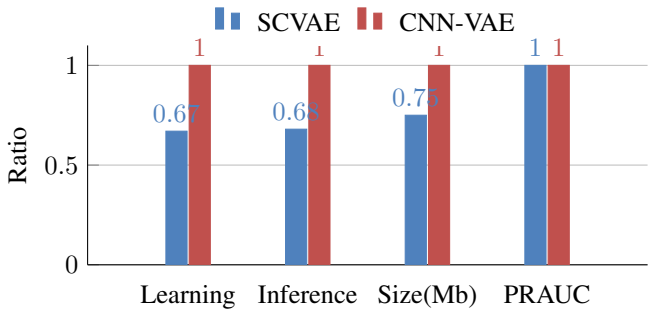
\captionsetup{font={footnotesize,rm},justification=centering,labelsep=period}

\captionsetup{font={footnotesize,sc},justification=centering,labelsep=period}%
\begin{table}[htbp]

\centering%
\renewcommand{\arraystretch}{1.3}
\begin{tabular}{C{1cm}C{1cm}C{1cm}C{1.2cm}C{1.2cm}}

\multicolumn{2}{l}{occupancy(16)} &               &                &                \\
\hline
\textit{Model} & \textit{Learning Time} & \textit{Inference Time} & \textit{Memory(Mb)} & \textit{PRAUC} \\
\hline
CNNVAE        & 3m 8sec       & 0.0088          & 16         & 98.9 \\
\textbf{SCVAE} & \textbf{2m 1sec} & \textbf{0.0060} & \textbf{12} & \textbf{99.2} \\
\hline
\end{tabular}
\caption{Comparison between SCVAE and CNN-VAE on occupancy(16) dataset}\label{model-size-inference-time}
\end{table}
\captionsetup{font={footnotesize,rm},justification=centering,labelsep=period}

Looking at Table \ref{model-size-inference-time} it can be seen that the learning time and inference time of the SCVAE model for the occupancy (16) dataset are reduced by approximately 1/3, compared to the CNN-VAE model. The memory size is also reduced from 16Mb to 12Mb, which is a decrease of about 1/4. In addition, it can be seen that there is little difference in the PRAUC value, meaning the model was successfully reduced. Looking at the Appendix, it can be seen that the larger the training data, the degree of reduction in the CNC data is greater.

\section{Conclusion \& Future Work}
In most manufacturing processes, labels of sensor data are often missing or inaccurate. The sensor data generated by the manufacturing industry has time series characteristics, and Edge Computing is attracting attention for real-time processing, security enhancement, and communication and computer infrastructure burden reduction.
In this paper, unsupervised anomaly detection is performed under an Edge Computing environment based on real manufacturing CNC data and UCI time series data.
Experiments show that neural network based anomaly detection models such as CNN-VAE and SCVAE outperform conventional machine learning algorithms in most cases.
Also, by comparing model size and inference time of SCVAE and CNN-VAE, it can be seen that SCVAE is a more suitable model for Edge Computing.

The metric Match-General is used in Experiment 4.B. 
However, it is not an objective metric and has many loopholes.
Future work will be to propose a metric that can compare the performance of anomaly detection algorithms on an objective basis in a label-free environment and perform comparison evaluations between more accurate algorithms. 


\section*{Acknowledgment}

This work was supported by Samsung Electronics Software Center, the BK21 Plus Program(Center for Sustainable and Innovative
Industrial Systems, Dept. of Industrial Engineering, Seoul National University) funded by the
Ministry of Education, Korea (No. 21A20130012638), the National Research Foundation(NRF)
grant funded by the Korea government(MSIP) (No. 2011-0030814), and the Institute for Industrial
Systems Innovation of SNU.



%
%
%

\bibliographystyle{IEEEtran}
\bibliography{bibtemplate_samples}

\begin{thebibliography}{1}
\providecommand{\url}[1]{#1}
\csname url@samestyle\endcsname
\providecommand{\newblock}{\relax}
\providecommand{\bibinfo}[2]{#2}
\providecommand{\BIBentrySTDinterwordspacing}{\spaceskip=0pt\relax}
\providecommand{\BIBentryALTinterwordstretchfactor}{4}
\providecommand{\BIBentryALTinterwordspacing}{\spaceskip=\fontdimen2\font plus
\BIBentryALTinterwordstretchfactor\fontdimen3\font minus
  \fontdimen4\font\relax}
\providecommand{\BIBforeignlanguage}[2]{{%
\expandafter\ifx\csname l@#1\endcsname\relax
\typeout{** WARNING: IEEEtran.bst: No hyphenation pattern has been}%
\typeout{** loaded for the language `#1'. Using the pattern for}%
\typeout{** the default language instead.}%
\else
\language=\csname l@#1\endcsname
\fi
#2}}
\providecommand{\BIBdecl}{\relax}
\BIBdecl

\bibitem{Firstman:2014:Approach}
J.~Firstman and B.~C. Secondman, ``\BIBforeignlanguage{english}{{An Approach
  for This and That}},'' \BIBforeignlanguage{english}{Journal of Something},
  vol. 125, 2014, pp. 1231--1238, {ISSN:} 9-9-9-9-9-9.

\bibitem{Firstman:2013:Ongoing}
J.~Firstman, B.~C. Secondman, and A.~C. Thirdman,
  ``\BIBforeignlanguage{english}{{Ongoing Research for This and That}},'' in
  \BIBforeignlanguage{english}{Proceedings of the 4$^{\textrm th}$
  International Conference on This and That (ICTT) July 12--14, 2013, Somecity,
  Somecountry}.\hskip 1em plus 0.5em minus 0.4em\relax Worldwide Publisher,
  Jul. 2013, pp. 1231--1238, {Edwards, C. U., Ed.}, {ISBN:} 9-9-9-9-9-9,
  {ISSN:} 99-99-99-99-99, {URL:} \url{http://www.someworldwideindex.org/}
  [accessed: 2014-01-01].

\bibitem{Webpage:2014:Somewebpage}
``\BIBforeignlanguage{english}{{Some Webpage}},'' 2014, {URL:}
  \url{http://www.somewebpage.org/} [accessed: 2014-01-02].

\bibitem{Bookman:2014:Somebook}
C.~U. Bookman, Ed., \BIBforeignlanguage{english}{{Some Book about This and
  That}}.\hskip 1em plus 0.5em minus 0.4em\relax The Somebook Publisher Inc.,
  Somewhere, Jan. 2014, {ISBN:} 978-99-99-99-99-9.

\bibitem{Chapman:2014:Starting}
R.~Chapman and S.~A. Chapman~Jr., \BIBforeignlanguage{english}{Starting with
  This and That}.\hskip 1em plus 0.5em minus 0.4em\relax The Somebook Publisher
  Inc., Somewhere, Jan. 2014, chapter 1, pp. 1--14, in Bookman, C. U., Some
  Book about This and That, {ISBN:} 978-99-99-99-99-9.

\end{thebibliography}


\begin{thebibliography}{10}
\providecommand{\url}[1]{#1}
\csname url@samestyle\endcsname
\providecommand{\newblock}{\relax}
\providecommand{\bibinfo}[2]{#2}
\providecommand{\BIBentrySTDinterwordspacing}{\spaceskip=0pt\relax}
\providecommand{\BIBentryALTinterwordstretchfactor}{4}
\providecommand{\BIBentryALTinterwordspacing}{\spaceskip=\fontdimen2\font plus
\BIBentryALTinterwordstretchfactor\fontdimen3\font minus
  \fontdimen4\font\relax}
\providecommand{\BIBforeignlanguage}[2]{{%
\expandafter\ifx\csname l@#1\endcsname\relax
\typeout{** WARNING: IEEEtran.bst: No hyphenation pattern has been}%
\typeout{** loaded for the language `#1'. Using the pattern for}%
\typeout{** the default language instead.}%
\else
\language=\csname l@#1\endcsname
\fi
#2}}
\providecommand{\BIBdecl}{\relax}
\BIBdecl

\bibitem{kingma2013auto}
D.~P. Kingma and M.~Welling, ``Auto-encoding variational bayes,'' arXiv
  preprint arXiv:1312.6114, 2013.

\bibitem{liu2008isolation}
F.~T. Liu, K.~M. Ting, and Z.-H. Zhou, ``Isolation forest,'' in Data Mining,
  2008. ICDM'08. Eighth IEEE International Conference on.\hskip 1em plus 0.5em
  minus 0.4em\relax IEEE, 2008, pp. 413--422.

\bibitem{breunig2000lof}
M.~M. Breunig, H.-P. Kriegel, R.~T. Ng, and J.~Sander, ``Lof: identifying
  density-based local outliers,'' in ACM sigmod record, vol.~29, no.~2.\hskip
  1em plus 0.5em minus 0.4em\relax ACM, 2000, pp. 93--104.

\bibitem{scholkopf2001estimating}
B.~Sch{\"o}lkopf, J.~C. Platt, J.~Shawe-Taylor, A.~J. Smola, and R.~C.
  Williamson, ``Estimating the support of a high-dimensional distribution,''
  Neural computation, vol.~13, no.~7, 2001, pp. 1443--1471.

\bibitem{rousseeuw1999fast}
P.~J. Rousseeuw and K.~V. Driessen, ``A fast algorithm for the minimum
  covariance determinant estimator,'' Technometrics, vol.~41, no.~3, 1999, pp.
  212--223.

\bibitem{an2015variational}
J.~An and S.~Cho, ``Variational autoencoder based anomaly detection using
  reconstruction probability,'' Technical Report, Tech. Rep., 2015.

\bibitem{goodfellow2014generative}
I.~Goodfellow, J.~Pouget-Abadie, M.~Mirza, B.~Xu, D.~Warde-Farley, S.~Ozair,
  A.~Courville, and Y.~Bengio, ``Generative adversarial nets,'' in Advances in
  neural information processing systems, 2014, pp. 2672--2680.

\bibitem{schlegl2017unsupervised}
T.~Schlegl, P.~Seeb{\"o}ck, S.~M. Waldstein, U.~Schmidt-Erfurth, and G.~Langs,
  ``Unsupervised anomaly detection with generative adversarial networks to
  guide marker discovery,'' in International Conference on Information
  Processing in Medical Imaging.\hskip 1em plus 0.5em minus 0.4em\relax
  Springer, 2017, pp. 146--157.

\bibitem{lee2017convolutional}
K.~B. Lee, S.~Cheon, and C.~O. Kim, ``A convolutional neural network for fault
  classification and diagnosis in semiconductor manufacturing processes,'' IEEE
  Transactions on Semiconductor Manufacturing, vol.~30, no.~2, 2017, pp.
  135--142.

\bibitem{han2015deep}
S.~Han, H.~Mao, and W.~J. Dally, ``Deep compression: Compressing deep neural
  networks with pruning, trained quantization and huffman coding,'' arXiv
  preprint arXiv:1510.00149, 2015.

\bibitem{iandola2016squeezenet}
F.~N. Iandola, S.~Han, M.~W. Moskewicz, K.~Ashraf, W.~J. Dally, and K.~Keutzer,
  ``Squeezenet: Alexnet-level accuracy with 50x fewer parameters and< 0.5 mb
  model size,'' arXiv preprint arXiv:1602.07360, 2016.

\bibitem{howard2017mobilenets}
A.~G. Howard, M.~Zhu, B.~Chen, D.~Kalenichenko, W.~Wang, T.~Weyand,
  M.~Andreetto, and H.~Adam, ``Mobilenets: Efficient convolutional neural
  networks for mobile vision applications,'' arXiv preprint arXiv:1704.04861,
  2017.

\bibitem{nair2010rectified}
V.~Nair and G.~E. Hinton, ``Rectified linear units improve restricted boltzmann
  machines,'' in Proceedings of the 27th international conference on machine
  learning (ICML-10), 2010, pp. 807--814.

\bibitem{glorot2010understanding}
X.~Glorot and Y.~Bengio, ``Understanding the difficulty of training deep
  feedforward neural networks,'' in Proceedings of the Thirteenth International
  Conference on Artificial Intelligence and Statistics, 2010, pp. 249--256.

\bibitem{ioffe2015batch}
S.~Ioffe and C.~Szegedy, ``Batch normalization: Accelerating deep network
  training by reducing internal covariate shift,'' in International Conference
  on Machine Learning, 2015, pp. 448--456.

\bibitem{kingma2014adam}
D.~Kingma and J.~Ba, ``Adam: A method for stochastic optimization,'' arXiv
  preprint arXiv:1412.6980, 2014.

\bibitem{Kun2008ozone}
\BIBentryALTinterwordspacing
X.~Y. Kun~Zhang, Wei~Fan, ``Ozone level detection data set,'' 2008. [Online].
  Available:
  \url{https://archive.ics.uci.edu/ml/datasets/Ozone+Level+Detection}
\BIBentrySTDinterwordspacing

\bibitem{Luis2016occupancy}
\BIBentryALTinterwordspacing
V.~F. Luis M.~Candanedo, ``{UCI} accurate occupancy detection of an office room
  from light, temperature, humidity and co2 measurements using statistical
  learning models,'' 2016. [Online]. Available:
  \url{https://archive.ics.uci.edu/ml/datasets/Occupancy+Detection}
\BIBentrySTDinterwordspacing

\end{thebibliography}

\clearpage

\appendix

\section{\\Title of Appendix A }

\captionsetup{font={footnotesize,sc},justification=centering,labelsep=period}%
\begin{table}[htbp]

\centering%
\renewcommand{\arraystretch}{1.3}
\begin{tabular}{C{0.6cm}C{0.4cm}C{1.4cm}C{1.2cm}C{0.8cm}C{0.8cm}C{1cm}}

\hline
\textit{Data} & \textit{Time Window} & \textit{Network} & \textit{Learning Time} & \textit{Inference Time} & \textit{Size(Mb)} & \textit{PRAUC} \\
\hline
ozone         & 4                    & SCVAE            & 1m 7s              & 0.014                  & 43                & 0.932          \\
              &                      & CNNVAE           & 1m 47s             & 0.021                  & 45                & 0.899          \\
              & 8                    & SCVAE            & 1m 33s             & 0.017                  & 85                & 0.773          \\
              &                      & CNNVAE           & 2m 53s             & 0.028                  & 128               & 0.841          \\
              & 16                   & SCVAE            & 2m 29s             & 0.022                  & 169               & 0.689          \\
              &                      & CNNVAE           & 5m 18s             & 0.043                  & 292               & 0.657          \\
              \hline
occupancy     & 4                    & SCVAE            & 1m 53s             & 0.005                  & 3.1               & 0.95           \\
\textit{}     &                      & CNNVAE           & 2m 55s             & 0.008                  & 5.4               & 0.943          \\
              & 8                    & SCVAE            & 2m                   & 0.005                  & 6                 & 0.977          \\
              &                      & CNNVAE           & 2m 54s             & 0.008                  & 8.9               & 0.923          \\
              & 16                   & SCVAE            & 2m 1s              & 0.006                   & 12                & 0.992          \\
\textit{}     &                      & CNNVAE           & 3m 8s              & 0.008                  & 16                & 0.989          \\
\hline
A        & 4                    & SCVAE            & 51m                  & 0.003                  & 19                & -              \\
              &                      & CNNVAE           & 1h 21m               & 0.005                   & 21                & -              \\
              & 8                    & SCVAE            & 1h 7m                & 0.004                  & 37                & -              \\
              &                      & CNNVAE           & 2h 10m               & 0.009                   & 55                & -              \\
              & 16                   & SCVAE            & 1h 47m               & 0.007                   & 73                & -              \\
              &                      & CNNVAE           & 4h 35m               & 0.015                  & 124               & -              \\
              \hline
B        & 4                    & SCVAE            & 1h 14m               & 0.004                  & 26                & -              \\
              &                      & CNNVAE           & 2h 2m                & 0.006                  & 28                & -              \\
              & 8                    & SCVAE            & 1h 45m               & 0.005                  & 51                & -              \\
              &                      & CNNVAE           & 3h 18m               & 0.011                   & 77                & -              \\
              & 16                   & SCVAE            & 2h 54m               & 0.009                   & 101               & -              \\
              &                      & CNNVAE           & 6h 5m                & 0.019                  & 173               & -              \\
              \hline
C         & 4                    & SCVAE            & 27m                  & 0.004                  & 26                & -              \\
              &                      & CNNVAE           & 44m                  & 0.006                  & 28                & -              \\
              & 8                    & SCVAE            & 38m                  & 0.005                  & 51                & -              \\
              &                      & CNNVAE           & 1h 11m               & 0.011                   & 77                & -              \\
              & 16                   & SCVAE            & 1h 5m                & 0.009                  & 101               & -              \\
              &                      & CNNVAE           & 2h 13m               & 0.019                  & 173               & -              \\
              \hline
D       & 4                    & SCVAE            & 2h 17m               & 0.003                  & 22                & -              \\
              &                      & CNNVAE           & 3h 48m               & 0.006                  & 25                & -              \\
              & 8                    & SCVAE            & 3h 9m                & 0.005                  & 44                & -              \\
              &                      & CNNVAE           & 5h 54m               & 0.009                  & 66                & -              \\
              & 16                   & SCVAE            & 5h                     & 0.007                  & 87                & -              \\
              &                      & CNNVAE           & 10h 20m              & 0.017                  & 148               & -              \\

\hline
\end{tabular}
\caption{Comparison between SCVAE and CNN-VAE}\label{scvaecnnvaecompare}
\end{table}
\captionsetup{font={footnotesize,rm},justification=centering,labelsep=period}

\end{document}